\documentclass{article}

\usepackage[final, nonatbib]{tackling_climate_workshop_style}
\usepackage{multicol}
\usepackage[utf8]{inputenc} % allow utf-8 input
\usepackage[T1]{fontenc}    % use 8-bit T1 fonts
\usepackage{hyperref}       % hyperlinks
\usepackage{url}            % simple URL typesetting
\usepackage{booktabs}       % professional-quality tables
\usepackage{amsfonts}       % blackboard math symbols
\usepackage{nicefrac}       % compact symbols for 1/2, etc.
\usepackage{amsmath}
\usepackage{graphicx}

\usepackage{subfig}

\usepackage{microtype}      % microtypography
\usepackage[square,numbers]{natbib}
\title{Evaluating Digital Tools for Sustainable Agriculture using Causal Inference}

\author{%
Ilias Tsoumas\thanks{Equal contribution.} $ ^{\text{ }1,2}$ \quad Georgios Giannarakis$^{*1}$ \quad Vasileios Sitokonstantinou$^1$ \\ \textbf{Alkiviadis Koukos}$^1$ \quad \textbf{Dimitra Loka}$^3$ \quad
\textbf{Nikolaos Bartsotas}$^1$ \\ \textbf{Charalampos Kontoes}$^1$ \quad \textbf{Ioannis Athanasiadis}$^2$\\
$^1$BEYOND Centre, IAASARS, National Observatory of Athens \\\quad $^2$Wageningen University and Research \\\quad $^3$Hellenic Agricultural Organization ELGO DIMITRA\\
\texttt{\{i.tsoumas,giannarakis,vsito,akoukos,nbartsotas,kontoes\}@noa.gr}\\
\texttt{\{ilias.tsoumas, ioannis.athanasiadis\}@wur.nl}\\
\texttt{dimitra.loka@elgo.gr}
}

\usepackage{floatrow} %ilias
\newfloatcommand{capbtabbox}{table}[][\FBwidth] %ilias

\usepackage{algorithm} % ADD ILIAS for algo
\usepackage{algorithmic} % ADD ILIAS for algo
\usepackage{nicefrac} % ADD ILIAS for algo
\usepackage{soul} % ilias st

\usepackage{blindtext}

\begin{document}

\maketitle

\begin{abstract}
In contrast to the rapid digitalization of several industries, agriculture suffers from low adoption of climate-smart farming tools. Even though AI-driven digital agriculture can offer high-performing predictive functionalities, it lacks tangible quantitative evidence on its benefits to the farmers. Field experiments can derive such evidence, but are often costly and time consuming.
% and hence limited in scope and scale of application.
To this end, we propose an observational causal inference framework for the empirical evaluation of the impact of digital tools on target farm performance indicators. 
% This way, we can increase farmers' trust via enhancing the transparency of the digital agriculture market, and in turn accelerate the adoption of technologies that aim to secure farmer income resilience and global agricultural sustainability against a changing climate.
This way, we can increase farmers' trust by enhancing the transparency of the digital agriculture market, and in turn accelerate the adoption of technologies that aim to increase productivity and secure a sustainable and resilient agriculture against a changing climate.
% As a case study, we designed and implemented a recommendation system for the optimal sowing time of cotton based on numerical weather predictions, which was used by a farmers' cooperative during the growing season of 2021. We then leverage agricultural knowledge, collected yield data, and environmental information to develop a causal graph of the farm system. Using the back-door criterion, we identify the impact of sowing recommendations on the yield and subsequently we estimate it using several methods.
As a case study, we perform an empirical evaluation of a recommendation system for optimal cotton sowing, which was used by a farmers' cooperative during the growing season of 2021. We leverage agricultural knowledge to develop a causal graph of the farm system, we use the back-door criterion to identify the impact of recommendations on the yield and subsequently estimate it using several methods on observational data.
% linear regression, matching, inverse propensity score weighting and meta-learners. 
The results show that a field sown according to our recommendations enjoyed a significant increase in yield ($12\%$ to $17\%$). 

% The effect estimates were robust, as indicated by the agreement among the estimation methods and four successful refutation tests. We argue that this approach can be implemented for decision support systems of other fields, extending their evaluation beyond a performance assessment of internal functionalities.
\end{abstract}

\section{Introduction}

The increasing global population and the changing climate are putting pressure on the agricultural sector, demanding the sustainable production of adequate quantities of nutritious food, feed and fiber. In this context, we need climate-smart agriculture \cite{lipper2014climate, fao} to optimize crop management with zero waste, enhance resilience, increase production and reduce emissions \cite{lynch2021agriculture}. Unfortunately, the agricultural sector experiences limited adoption of pertinent smart farming technologies \cite{gabriel2022adoption} that could drive the required sustainable production. This might seem odd at first sight, given the recent surge of sophisticated digital tools that utilize Artificial Intelligence (AI) and big Earth data \cite{sharma2020machine}; yet farmers are skeptical about their effectiveness as most lack quantitative evidence on their benefits \cite{lowenberg2019setting, lioutas2021digitalization}.
% to their revenues and daily work
% This might seem odd at first sight, given the surge of sophisticated digital tools that utilize Artificial Intelligence (AI) techniques and combine remote sensing data with data from Internet of Things (IoT) sensors to offer agricultural information of great detail \cite{sharma2020machine}. Yet farmers are skeptical about the effectiveness and actual contribution of these tools to their revenues and daily work \cite{lowenberg2019setting, lioutas2021digitalization}.
Traditionally, quantifying the impact of a service would require the design and execution of a randomized experiment \cite{boruch1997randomized}.  Nevertheless, field experiments for the evaluation of digital agriculture tools are seldom done since they are inflexible, requiring follow-up experiments for any changes in the product, but also costly and time-consuming \cite{vaessen2010challenges}. 
% In addition, field experiments can jeopardize the crops and hence the farmers' livelihood; and if potential damages are not covered, no prudent farmer would want to participate. 
% As a result, for lack of a better alternative, the providers of digital agriculture tools often resort to unproven promises that unavoidably create customer mistrust. 
Thus, an observational causal inference framework \cite{pearl2009causality} can fill this gap by emulating the experiment we would have liked to run \cite{hernan2016using}. Causal inference with observational data has been the subject of recent work across diverse disciplines, including ecology \cite{arif2022utilizing}, public policy \cite{fougere2019causal, fougere2021policy}, and Earth sciences \cite{massmann2021causal, perez2018causal, runge2019inferring}. In agriculture, it has been used to identify and estimate the effect of agricultural practices on various agro-environmental metrics \cite{qian2016applying, deines2019satellites, giannarakis2022towards}.
% Causal inference offers a dependable middle-ground between the hard-earned knowledge from experiments and the unreliable conclusions from naive observations or predictions \cite{hernan2010causal}. This is a trade-off of particular relevance to the evaluation of digital agriculture systems. The stakes are not as high as in e.g., drug testing; hence, strict experimental protocols are not needed.
% Using causal inference to test digital agriculture can provide reliable insights of superior socioeconomic impact than of those inferred by naive descriptive studies, i.e., transparent benefits for the farmers, increased reliability, honest pricing. 
According to Adelman (1992), the comprehensive evaluation of decision support systems has three facets: i) the subjective 
% evaluation that assesses the system from the perspective of the end-user
, ii) the technical 
% evaluation that assesses the performance of the system's internal functionalities 
and iii) the empirical evaluation 
% that experimentally assesses the impact of the system
\cite{adelman1992evaluating}. While subjective and technical evaluation have been sufficiently practiced \cite{zhai2020decision, schroder2011setting},
% Subjective evaluation has been widely practiced for decision support  \cite{zhai2020decision}  and recommender systems \cite{pu2012evaluating}, retrieving user feedback (e.g., on usability, accessibility etc.) through surveys and questionnaires. From the perspective of the AI-based expert systems, the technical evaluation is based on predictive, classification and ranking accuracy metrics \cite{schroder2011setting}. Technical evaluation metrics extending beyond accuracy have been proposed within the context of recommender systems (e.g., coverage, serendipity) \cite{ge2010beyond}. These metrics try to quantitatively measure the recommendation quality as perceived by the user, connecting the technical and subjective evaluation concepts.
the empirical evaluation methods, and in particular with regards to the impact assessment of digital agriculture tools, have been seldom employed. 
% Compared to its technical and subjective counterparts, empirical evaluation requires a fundamentally different approach endowed with the ability of causal reasoning. 
% From the perspective of agricultural economics, tangential questions have been studied \cite{muller1974sources, chavas1984information, roberts2009estimating, schimmelpfennig2016sequential, mcfadden2022information}, but without approaching the question from a farm system standpoint, hence not leveraging available structural knowledge and reaping its benefits \cite{cinelli_crash_2020}.
% A study with similar vision to this one \cite{mcfadden2022information} tries to quantify the efficiency of yield and soil maps using survey data and the econometrical approach of stochastic frontier analysis. 
Thus, we propose a framework for the empirical evaluation of digital agriculture recommendations with causal inference.
% wish to demonstrate how causal inference can be used for the empirical evaluation of digital agriculture recommendations. 
In this context, we evaluate a recommendation system for the optimal sowing of cotton, given sowing time is of great importance for arable crops. Mistimed sowing can lead to suboptimal plant emergence and adversely affect the crop yield \cite{huang2016different,bradow2010germination,bauer1998planting,nielsen2002delayed,richards2022impact}. 

To the best of our knowledge, there are no works that evaluate the effectiveness of any type of decision support or recommendation system in the agricultural sector through causal reasoning and 
beyond their predictive accuracy \cite{luma2020causal, pasquel2022review}. The contributions of this work are summarized as follows: i) the design of the first empirical evaluation framework for digital agriculture based on causal inference; ii) the implementation of it to assess the impact of a recommendation system, which was operationally used in a real-world case study;
% based on numerical weather predictions and agricultural knowledge.
% Prior to our work, the farmers of the agricultural cooperative where the system was tested did not have an equivalent decision support tool;
% Also, this is one of only a handful of optimal cotton sowing tools in the world and the one with the highest spatial resolution (2 x 2 km)
iii) the identification of the causal effect of sowing recommendations on yield, its subsequent estimation, and the evaluation of estimates using refutation tests. 

% Our system increased the yield of the farmers that followed the recommendations by a factor that ranged from 12\% to 17\%, depending on the estimation method used.
 
\section{Case Study}
% \subsection{Data \& Problem Formulation}
% \paragraph{Agricultural Recommendation System}
In this work, we implement the empirical evaluation of a knowledge-based recommendation system \cite{aggarwal2016knowledge} for optimal cotton sowing, which aims to make farmers' production, and hence their profit, resilient against climate change. The recommendations are based on satisfying specific environmental conditions, as retrieved from the related literature, which would ensure successful cotton planting. The system is operationally deployed using high resolution weather forecasts. A.2 of the Appendix contains the design, implementation, algorithmic presentation and the technical evaluation of the system. We provided the recommendations in the form of daily maps, indicating unfavorable and favorable conditions, over the fields of the participating farmers. The sowing recommendation maps were served through the website of their cooperative, which farmers visited on a daily basis during the growing season of 2021. The cooperative collected and provided the required data for each field (i.e., geo-referenced boundaries, sowing \& harvest date, seed variety, yield). We then combined this data with publicly available observations from heterogeneous sources (i.e., Sentinel-2 images, climate variables, soil maps) to engineer an observational dataset that enables the causal analysis.

% The cooperative collected and provided  for each field: its geo-referenced boundaries, the sowing date, the seed variety, the harvest date, the precise final yield, and for a subset of the fields the yield of the previous year. We then combined this data with publicly available observations from heterogeneous sources, such as satellites (Sentinel-2), weather stations and GIS maps, to engineer an observational dataset that enables a causal analysis for studying the impact of the recommendation system on the yield.

\section{Causal Evaluation Framework}

\paragraph{Notation \& Terminology.}

We encode the farm system in the form of a Directed Acyclic Graph (DAG) $G \equiv (V,E)$ where $V$ is a set of vertices consisting of all relevant variables, and $E$ is a set of directed edges connecting them \cite{pearl2009causality}. The directed edge $A \rightarrow B$ indicates causation from $A$ to $B$, in the sense that changing the value of $A$ and holding everything else constant will change the value of $B$. We are using Pearl's $do$-operator to describe interventions, with $\mathbb{P}(Y=y|do(T=t))$ denoting the probability that $Y = y$ given that we intervene on the system by setting the value of $T$ to $t$. 
% Following popular terminology, w
We name the variable $T$, of which we aim to estimate the effect, as \textit{treatment} and the variable $Y$, which we want to quantify the impact of $T$ on, as \textit{outcome}. The parents of a node are its \textit{direct causes}, while a parent of both the treatment and outcome is referred to as a \textit{common cause} or \textit{confounder}. Our end goal is to account for exactly the variables $Z \subseteq V$ that will allow us to estimate the Average Treatment Effect (ATE) of the treatment on outcome, as shown in Eq. (\ref{eq:ate}).
\begin{equation}\label{eq:ate}
    \text{ATE} = \mathbb{E}\big[Y|do(T=1)\big] - \mathbb{E}\big[Y|do(T=0)\big]
\end{equation}

\paragraph{Problem Formulation \& Causal Graph.}

\begin{figure}[!ht]
\begin{floatrow}
\ffigbox{%
\hspace*{-0.5cm}
  \includegraphics[scale=0.18]{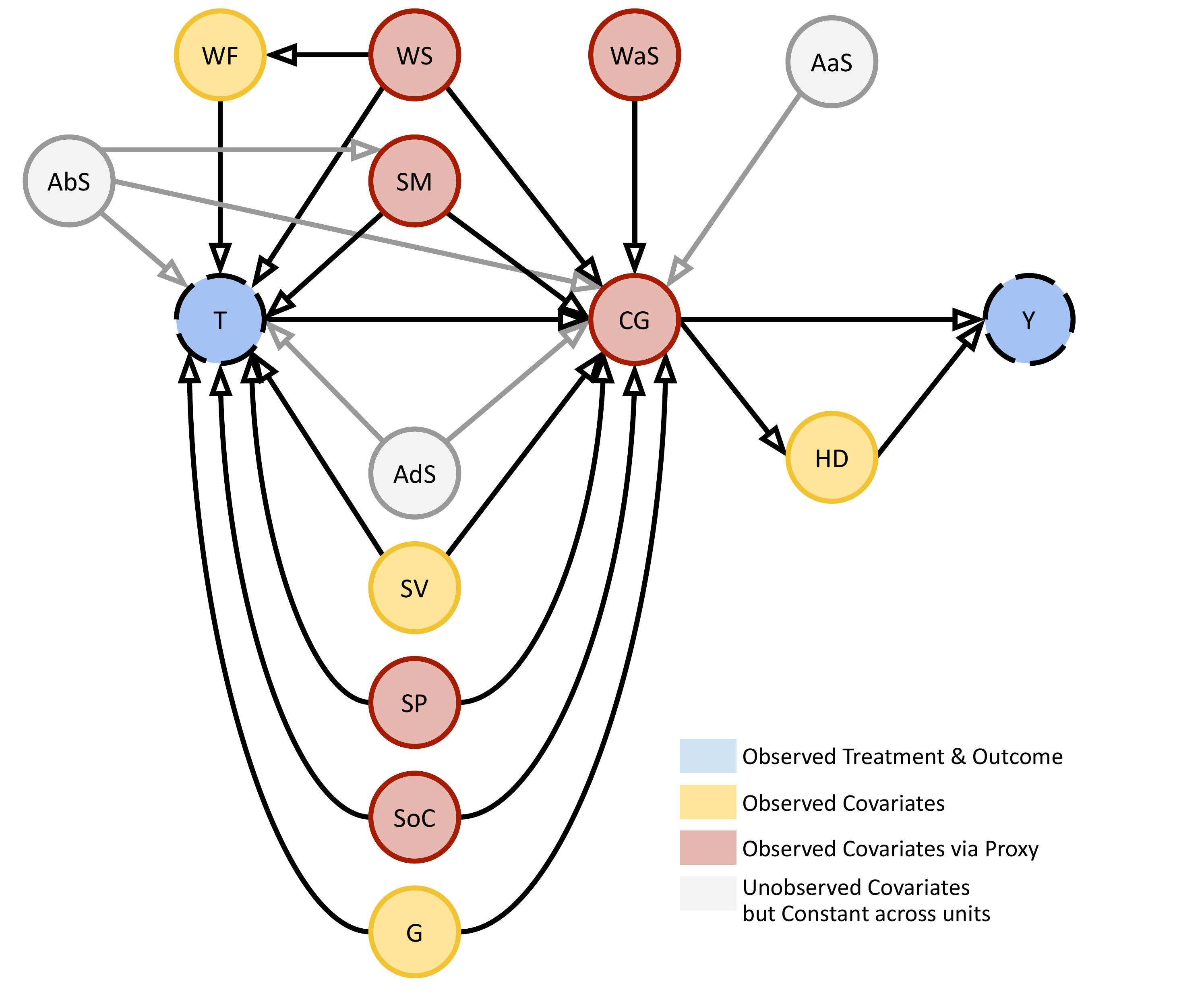}
}{%
    \caption{Graph of the farm system. \label{fig:causal-graph}}%
}
\capbtabbox{%
% \begin{table}[!ht]
% \small
% \centering
% \resizebox{\columnwidth}{!}{%
\hspace*{1cm}
\begin{tabular}{lll}
\toprule
\textbf{Id} & \textbf{Variable Description}                    \\ \midrule
T   & Treatment       
% T   & Treatment & Recommender System    
\\
% Y   & Outcome (Yield) & Agriculture Cooperative \\
WF  & Weather forecast                             \\
WS  & Weather on sowing day          \\
WaS & Weather after sowing                           \\
CG  & Crop Growth                                       \\
SM  & Soil Moisture on sowing           \\
SP  & Topsoil physical properties   \\
SoC & Topsoil organic carbon       \\
SV  & Seed Variety                 \\
G   & Geometry of field                             \\
AdS & Practices during sowing      \\
AbS & Practices before sowing       \\
AaS & Practices after sowing        \\
% LS  & Length of Season             & Agriculture Cooperative \\
HD  & Harvest Date                  \\
Y   & Outcome (Yield)               \\ \bottomrule
\end{tabular}%
% \caption{Farm system variable identifier, description and source.}
% \label{tab:variables}
% % }
% \end{table}
}{%
  \caption{Variables identifier and description.}%
  \label{tab:variables}
}
\end{floatrow}
\end{figure}

We thus aim to develop a causal graph $G$ whose vertices $V$ capture the relevant actors of the system we study, and edges $E$ indicate their relationships. 
% Recall that the end goal is the impact assessment of the decision support system's recommendations. 
The system recommendations should be part of the graph, along with cotton yield and the agro-environmental conditions that interfere in this physical process. Because the end goal is the evaluation of the recommendation system and its actual impact on yield, we designate as \textit{treated} the fields that farmers sowed on a day that was seen as favorable by the system, 
% i.e., the corresponding 10-day ART forecast satisfied the appropriate conditions,
and as \textit{control} the fields that were sown on a non-favorable day. 
% Because the system outputs $4$ levels of recommendation ranging from $0$ (bad) to $3$ (good), 
We define a day as favorable when all environmental conditions are satisfied. 
% i.e., when the system outputs the highest recommendation value that is $3$. 
% A day is then defined as non-favorable when the system outputs any other recommendation value. 
Binarizing the treatment in that way allows for greater flexibility in estimator selection and easier interpretation.
Beyond the recommendation system, multiple factors influence the decision to sow or not. This is precisely the challenge we aim to address by employing a graphical analysis and explicitly modeling the farm system structure. The ATE we aim to estimate captures the difference between what the average yield would have been if we intervened and forced farmers to follow the recommendation by sowing on a favorable day, and the average yield if we forced them to defy the recommendation by sowing on an unfavorable day. 
% Such an estimand is of primary significance for the farmers, but also for proving the reliability and therefore accelerating the adoption of smart farming tools.
% Because the recommendation system works by applying domain knowledge on weather forecasts, this quantity essentially coincides with the causal effect of weather on yield from sowing day and up to 10 days after. By quantifying this effect we evaluate the recommendation system's impact on yield, since farmers had no access to information of this precision and simplicity before, as emphasized by the fact that sowing decisions were basically a result of guesswork. 
Given that 
% treatment $T$ was defined as sowing in conditions the system considered favorable $(T=1)$ against non-favorable $(T=0)$ and that 
confounding factors are controlled for, we henceforth refer to the ATE as the \textit{(average) causal effect of following the recommendation} in the sense described above. Figure \ref{fig:causal-graph} displays the final causal graph $G$. We note that, in reality, it is impossible to account for all factors interacting in the system in order to claim that the estimated effect will not contain any bias. However, because the selection of variables is deeply rooted on well-understood agro-environmental interactions (detailed analysis of graph building in A.2 of Appendix), bias is expected to be minimized, in the sense that no important interactions are left unaccounted for. Furthermore, we extensively test the reliability of effect estimates through multiple refutation checks.

\paragraph{Identify, Estimate Effect \& Refute Estimate.}

Because the calculation of causal effects requires access to counterfactual values that are by definition not observed \cite{holland1986statistics}, observational methods rely on identification techniques and assumptions that aim at reducing causal estimands such as $\mathbb{P}(Y=y|do(T=t))$ to statistical ones, such as $\mathbb{P}(Y=y|T=t)$. The back-door criterion is a popular identification method that solely relies on a graphical test to infer whether adjusting for a set of graph nodes $Z\subseteq V$ is sufficient for identifying $\mathbb{P}(Y=y|do(T=t))$ from observational data.
% Formally, a set of variables $Z$ satisfies the back-door criterion relative to an ordered pair of variables $(T, Y)$ in a DAG $G$, if no node in $Z$ is a descendant of $T$ and $Z$ blocks every path between $T$ and $Y$ that contains an arrow into $T$.
%Given a DAG $G$, modern causal inference software automatically searches for $Z$.
% After (if) we have obtained a back-door adjustment set to condition on, we can proceed with estimating the ATE of interest. The back-door criterion already provides a formula for the interventional distribution. Given a set of variables $Z$ satisfying the back-door criterion we can identify the causal effect of $T$ on $Y$ as $\mathbb{P}(y|do(t)) = \sum_z \mathbb{P}(y|t, z)\mathbb{P}(z)$. 
After (if) we have obtained an adjustment set of variables $Z$ satisfying the back-door criterion we can identify the causal effect of $T$ on $Y$ as $\mathbb{P}(y|do(t)) = \sum_z \mathbb{P}(y|t, z)\mathbb{P}(z)$. 
% The ATE then reduces down to:

% \begin{equation}
%     \mathbb{P}(y|do(t)) = \sum_z \mathbb{P}(y|t, z)\mathbb{P}(z)
% \end{equation}

% \begin{equation}
%     \text{ATE} = \mathbb{E}_Z \big[\mathbb{E}[Y|T=1,Z] - \mathbb{E}[Y|T=0,Z] \big]
% \end{equation}

In our study, ATE estimation is done with several methods of varying complexity. 
% To check covariate balance and as a method prerequisite, we model the propensity scores $\mathbb{P}(T=1|Z=z)$, i.e., the probability of receiving treatment given features \cite{rosenbaum1983central}. 
Linear regression and distance matching are selected as baseline estimation methods. The popular Inverse Propensity Score (IPS) weigthing is also used \cite{stuart2010matching}. We finally apply modern machine learning methods, i.e., the baseline T-learner and the state-of-the-art X-learner \cite{kunzel_metalearners_2019}.
% \subsubsection{Refutation Methods.}

% One of the biggest challenges in causal inference pertains to model evaluation. 
Given the fact that ground truth estimates are not observed, we resort to performing robustness checks and sensitivity analyses of estimates, in line with recent research \cite{sharma2020dowhy, cinelli2020making}. We perform the following tests: i) Placebo treatment, where the treatment is randomly permuted and the estimated effect is expected to drop to $0$; ii) Random Common Cause (RCC), where a random confounder is added to the dataset and the estimate is expected to remain unchanged; iii) Random Subset Removal (RSR), where a subset of data is randomly selected and removed and the effect is expected to remain the same; iv) Unobserved Common Cause (UCC), where an unobserved confounder acts on the treatment and outcome without being added to the dataset, and the estimates should remain relatively stable. 

% The Placebo, RCC and RSR tests are bootstrapped to generate confidence intervals and p-values \cite{diciccio1996bootstrap}. The UCC returns a heatmap of new ATE estimates depending on the strength of unobserved confounding.

\section{Experiments and Results}

\begin{table}[!ht]
\centering
\resizebox{\textwidth}{!}{
\begin{tabular}{cccccccccccc}
\toprule
\multicolumn{4}{c}{\textbf{Causal Effect Estimation}} & \multicolumn{8}{c}{\textbf{Refutations}}                                              \\ \cmidrule(l){5-12} 
\multicolumn{4}{c}{} &
  \multicolumn{2}{c}{\textbf{Placebo}} &
  \multicolumn{2}{c}{\textbf{RCC}} &
  \multicolumn{2}{c}{\textbf{UCC}} &
  \multicolumn{2}{c}{\textbf{RRS}} \\ \midrule
\textbf{Method} & \textbf{ATE} & \textbf{CI} &  \textbf{p-value} 
& \textbf{Effect*} & \textbf{p-value} & \textbf{Effect*} &
  \textbf{p-value} &   \multicolumn{2}{c}{\textbf{Effect*}} &
  \textbf{Effect*} & \textbf{p-value} \\ 
  \midrule
\textbf{Linear Regression}   & 546   & (211, 880)   & 0.0015  & -25.74 & 0.39 & 546 & 0.49  & \multicolumn{2}{c}{85} & 543 & 0.45 \\
\textbf{Matching}            & 448   & (186, 760)   & 0.0060   & 50.82 & 0.39 & 432  & 0.40 & \multicolumn{2}{c}{116} & 438 & 0.48 \\
\textbf{IPS weighting} &  471 &  (138, 816) &  0.0010 &  38.82 &  0.40 &  470 &  0.40 &  \multicolumn{2}{c}{113} &  462 &  0.45 \\
\textbf{T-Learner (RF)}      & 372   & (215, 528)   & 0.0240 & 9.26  & 0.49 & 373 & 0.46 & \multicolumn{2}{c}{-} & 353 & 0.42 \\
\textbf{X-Learner (RF)}      & 437   & (300, 574)   & 0.0050 & 5.10   & 0.50 & 430 & 0.37 & \multicolumn{2}{c}{-} & 409    & 0.36 \\ \bottomrule
\end{tabular}
\caption{ATE point estimates, $95\%$ confidence intervals and p-values. Refutation tests fail if their p-value is less than 0.05. Numbers are in cotton kg/ha.}
\label{tab:results}
}
\end{table}
The sowing period lasted from early April to early May, the harvest took place in September, and yields ranged from $1,250$ to $6,960$ kg/ha.
%We dropped two fields that declared $0$ yield and were not harvested.
The dataset consists of $171$ fields ($51$ treated and $120$ control). 
% Two fields declared $0$ yield since they did not bear profitable amount of cotton lint. 
% Variables that registered intra-field values (NDVI, NDWI) were averaged at the field-level. 
% For the experiments, we are using the popular doWhy \cite{sharma2020dowhy} and Causal ML \cite{chen2020causalml} Python libraries.
Applying the back-door criterion on graph $G$ (Figure \ref{fig:causal-graph}), the following adjustment set of nodes $Z = \{\textsc{ws\textsubscript{min, max}, soc, sm, g, sp\textsubscript{silt, clay, sand}, abs, ads, sv\textsubscript{1-13}} \}$ was found sufficient for identifying the ATE. Variables in $Z$ are numerical, including the one-hot encoded vectors of the categorical \textsc{SV}\textsubscript{1-13} variable of variety. $AbS$ and $AdS$ are constant and thus excluded from estimation methods. 

Table \ref{tab:results} show the results of the ATE estimation per method, alongside $95\%$ confidence intervals and p-values. Besides Linear Regression, 
% do not provide confidence intervals by default. For matching, IPS, and meta-learners 
confidence intervals and the resulting p-values are bootstrapped. Both the T-learner and X-learner use a Random Forest for modeling the outcome $Y$. All methods detect a significant ATE at $95\%$ confidence level, with point estimates ranging from $372$ to $546$ kilograms of cotton per hectare. For context, the average observed yield is $3,145$ kg/ha. We thus infer that the causal effect of following the sowing recommendation on yield is significantly positive, driving a yield increase ranging from $12\%$ to $17\%$.
% Of central importance are the refutation tests we run after having estimated the recommendation impact. Table \ref{tab:results} features analytic results for all method / refutation test combinations. All estimation methods are robust against performing the following data manipulations and re-estimating the ATE: randomly permuting the treatment (Placebo test), adding a confounder (RCC test), sampling a subset of data (RSR test) and creating unobserved confounding (UCC test). Specifically, Placebo ATE estimates do not differ significantly from $0$, while RCC and RSR estimates do not differ significantly from the already obtained ATE. For the UCC test, the mean ATE estimates are reduced yet remain positive, despite unobserved confounding of significant magnitude. Confidence intervals and p-values are bootstrapped ($1000$ iterations).
Furthermore, Table \ref{tab:results}  illustrates that estimation methods are robust against refutation tests. Specifically, Placebo ATE estimates do not differ significantly from $0$, while RCC and RSR estimates do not differ significantly from the already obtained ATE. For the UCC test, the mean ATE estimates are reduced yet remain positive, despite unobserved confounding of significant magnitude (more details in part A.3 of the Appendix). 
The results indicate that the recommendation system's advice drove a net increase in yield that was both statistically significant and robust. Therefore, farmers are equiped with a provably valuable tool that optimizes the chances of a successful growing season with higher production, and lowers the likelihood of resorting to expensive actions and wasting resources, e.g., replanting the field.

\section{Conclusion}

In this study, we design, implement, and test a digital agriculture recommendation system for the optimal sowing of cotton. Using the collected data and leveraging domain knowledge, we evaluate the impact of system recommendations on yield. To do so, we utilize and propose causal inference as an ideal tool for empirically evaluating decision support systems. This idea can be upscaled to other digital agriculture tools as well as to different fields with well-established domain knowledge. This paradigm is in principle different to decision support systems that frequently use black-box algorithms to predict variables of interest, but are oblivious to the evaluation of their own impact.
% In that sense, this work comes to the defence of the farmer, by introducing an AI framework for elaborating on the assumptions, reliability, and impact of a system before discussing service fees.
In that sense, this work comes to empower the farmer towards resilient agriculture, by introducing an AI framework for elaborating on the assumptions, reliability, and impact of a system that promises green and climate-smart advice.

\begin{ack}
This work has been supported by the EU Horizon 2020 Research and Innovation program through the following projects: EIFFEL (grant agreement No. 101003518), CALLISTO (grant agreement No. 101004152), e-shape (grant agreement No. 820852), EXCELSIOR (grant agreement No. 857510). It was also supported by the MICROSERVICES project (2019-2020 BiodivERsA joint call, under the BiodivClim ERA-Net COFUND programme, and with the GSRI, Greece - No. T12ERA5-00075).

\end{ack}

% \section*{References}
\medskip

\bibliography{ref}

% References follow the acknowledgments. Use unnumbered first-level heading for
% the references. Any choice of citation style is acceptable as long as you are
% consistent. It is permissible to reduce the font size to \verb+small+ (9 point)
% when listing the references.
% {\bf Note that the Reference section does not count towards the pages of content that are allowed; 4 pages for Papers track and 3 pages for Proposals track.}
% \medskip

% \small

% [1] Alexander, J.A.\ \& Mozer, M.C.\ (1995) Template-based algorithms for
% connectionist rule extraction. In G.\ Tesauro, D.S.\ Touretzky and T.K.\ Leen
% (eds.), {\it Advances in Neural Information Processing Systems 7},
% pp.\ 609--616. Cambridge, MA: MIT Press.

% [2] Bower, J.M.\ \& Beeman, D.\ (1995) {\it The Book of GENESIS: Exploring
%   Realistic Neural Models with the GEneral NEural SImulation System.}  New York:
% TELOS/Springer--Verlag.

% [3] Hasselmo, M.E., Schnell, E.\ \& Barkai, E.\ (1995) Dynamics of learning and
% recall at excitatory recurrent synapses and cholinergic modulation in rat
% hippocampal region CA3. {\it Journal of Neuroscience} {\bf 15}(7):5249-5262.
\appendix{
\section{Supplementary Material}
\subsection{Agricultural Recommendation System}

In this work, we design, implement and evaluate a knowledge-based recommendation system \cite{aggarwal2016knowledge} for optimal cotton sowing. The recommendations are based on satisfying specific environmental conditions, as retrieved from the related literature, which would ensure
successful cotton planting. The system is operationally deployed using high resolution weather forecasts. Sec. 1 of the Appendix contains an algorithmic presentation of the system.

According to literature, the minimum daily-mean soil temperature for cotton germination is $16^\circ$C \cite{bradow2010germination}. Soil or ambient temperatures lower than $10^\circ$C result in less vigorous and malformed seedlings \cite{boman2005soil}. As a general rule for cotton, agronomists recommend daily-mean soil temperatures higher than $18^\circ$C for at least 10 days after sowing and daily-maximum ambient temperatures higher than $26^\circ$C for at least 5 days after sowing. We summarize the conditions for optimal cotton sowing in Table~\ref{fig:ag_rules} \cite{freeland2006agrometeorology, boman2005soil}. Using these conditions and Numerical Weather Predictions (NWP) we implement a recommendation system that advises on whether any given day is a good day to sow or not.

% These rules were dictated the corresponding variables guide the numerical weather prediction (NWP) model on choosing the variables needed in order to construct a sowing recommendation for a given day.

% \usepackage{booktabs}
% \usepackage{graphicx}
\begin{table}[!ht]
\centering
% \resizebox{\columnwidth}{!}{%
\begin{tabular}{@{}lllll@{}}
\toprule
\textbf{\begin{tabular}[l]{@{}l@{}}Type of \\ Temperature\end{tabular}} &
  \textbf{Statistic} &
  \textbf{Condition} &
%   \textbf{\begin{tabular}[c]{@{}l@{}}Time Window\\ (next days)\end{tabular}} &
  \textbf{\begin{tabular}[c]{@{}l@{}}Condition\\ Priority\end{tabular}} \\ \midrule
soil (0-10 cm)  & mean & \textgreater{}$18^\circ$C    &  optimum   \\
ambient (2 m) & max  & \textgreater{}$26^\circ$C & optimum   \\
soil (0-10 cm)   & mean & \textgreater{}$15.56^\circ$C & mandatory \\
soil (0-10 cm)   & min  & \textgreater{}$10^\circ$C   & mandatory \\
ambient (2 m) & min  & \textgreater{}$10^\circ$C   & mandatory \\ \bottomrule
\end{tabular}%
% }
\caption{Optimal conditions for sowing cotton. All conditions refer to the period from sowing day to 5 days after, except the first soil condition that refers to 10 days after. \label{fig:ag_rules}}
\end{table}

% \subsection{Employment of NWP}
% Evaluation of WRF, GFS, ARTIFICIAL10
% Contrary to the vast availability of high-resolution satellite remote sensing data,

Open-access high-resolution NWP forecasts are rarely available. For this reason, we implement the WRF-ARW model \cite{skamarock2019description} with a grid resolution of 2 km. This enables us to reach a high spatio-temporal resolution for parameters that are crucial during the cotton seeding period, namely the soil and ambient temperature that are retrieved in hourly rate for the forthcoming 2.5 days. 
% The Noah Land-Surface submodel \cite{niu2011community} calculates online the energy flux between the atmosphere and land to provide soil temperature at four depths (0-0.1 cm, 0.1-1 cm, 1-10 cm, 10-40 cm). 
Ideally, 10-day predictions at a 2 km spatial resolution should be available every morning, as it is required by the conditions in Table \ref{fig:ag_rules}. However, this would demand an enormous amount of computational power. To simulate the desired data, we combine the 2.5-day high resolution forecasts with the GFS \cite{cisl_rda_ds084.1} 15-day forecasts that are given on a 0.25 degrees (roughly 25 km) spatial resolution.
% The ideal would be to have a 2-km, 10-day simulation available each morning, 

\begin{equation} \label{eq:a_gfs}
a_{i}= \frac{GFS_{day=i}}{GFS_{day=1}}, i\in \{3, ..., 10\}
\end{equation}

\begin{equation} \label{eq:art}
ART_{j}=\left\{
\begin{array}{ll}
WRF_{day=j} &, j\in \left \{ 1,2 \right \} \\ 
WRF_{day=1}\cdot  a_{j} &, j\in \{3, ..., 10\}
\end{array}
\right.
\end{equation}

% \begin{equation} \label{eq:a_gfs}
% a_{i}= 1-\frac{GFS_{day=1}}{GFS_{day=i}}, i\in \left [ 3,10 \right ]
% \end{equation}

% \begin{equation} \label{eq:art}
% ART_{j}=\left\{
% \begin{array}{ll}
% WRF_{day=j} &, j\in \left \{ 1,2 \right \} \\ 
% WRF_{day=1}\times \left ( 1+a_{j} \right ) &, j\in\left [ 3,10 \right ] 
% \end{array}
% \right.
% \end{equation}

Eq. (\ref{eq:a_gfs}) shows how we extract the 10-day weather trend factor using GFS forecasts. We calculate the percentage change between each forecast (for $day=3$ to $day=10$) and the corresponding next day ($day=1$) forecast.
Eq. (\ref{eq:art}) shows how we produce the artificial (ART) 10-day forecasts at 2 km spatial resolution. We keep the original WRF forecasts for the next two days and for the rest we apply the respective 10-day trend factor to the next day WRF forecast.

% \begin{algorithm}[!h]
% \caption{Optimal sowing recommendation system}
% \label{alg:algorithm}
% \begin{flushleft}\textbf{Input}: WRF, GFS, AoI\\\end{flushleft}
% \begin{flushleft}\textbf{Parameter}: Gridpoints coupled with weather variables\\\end{flushleft}
% \begin{flushleft}\textbf{Output}: Map of recommendations\end{flushleft}
% \begin{algorithmic}[1] %[1] enables line numbers
% \FOR{$day$ and $gpoint$ of $GFS$ where $3\leq day\leq 10$}
% \STATE $a_{gpoint,day} \leftarrow \nicefrac{GFS_{day=[3,10]}}{GFS_{1}}$.
% \ENDFOR
% \FOR{$gpoint_W$ in $WRF$}
% \FOR{$gpoint_G$ in $GFS$}
% \STATE $npoint \leftarrow nneighbour(gpoint_W,gpoint_G)$.
% \FOR{$day$ in \{1, ..., 10\}}
% \IF {$day<3$}
% \STATE $ART_{gpoint_W,day} \leftarrow WRF_{gpoint_W,day}$.
% \ELSE
% \STATE $ART_{gpoint_W,day} \leftarrow WRF_{gpoint_W,1} \cdot a_{npoint,day}$.
% \ENDIF
% \ENDFOR
% \ENDFOR
% \ENDFOR
% \FORALL{$day$ in each $gpoint$ in $ART$}
% % \STATE $opt1,opt2,mand1,mand2,mand3 \leftarrow False$\;.
% \IF {$all$ $weather$ $rules == True$}
% \STATE $sowing_{gpoint} \leftarrow True$.
% \ELSE
% \STATE $sowing_{gpoint} \leftarrow False$.
% \ENDIF
% \ENDFOR
% \FOR{$lat,lon$ in $AoI$}
% \STATE $map \leftarrow 
% nearest$ $neighbour(AoI,sowing)$.
% \ENDFOR
% \STATE \textbf{return} colored map of sowing recommendations
% \end{algorithmic}
% \end{algorithm}

\begin{figure}[!ht]
  \centering
  \includegraphics[scale=0.25]{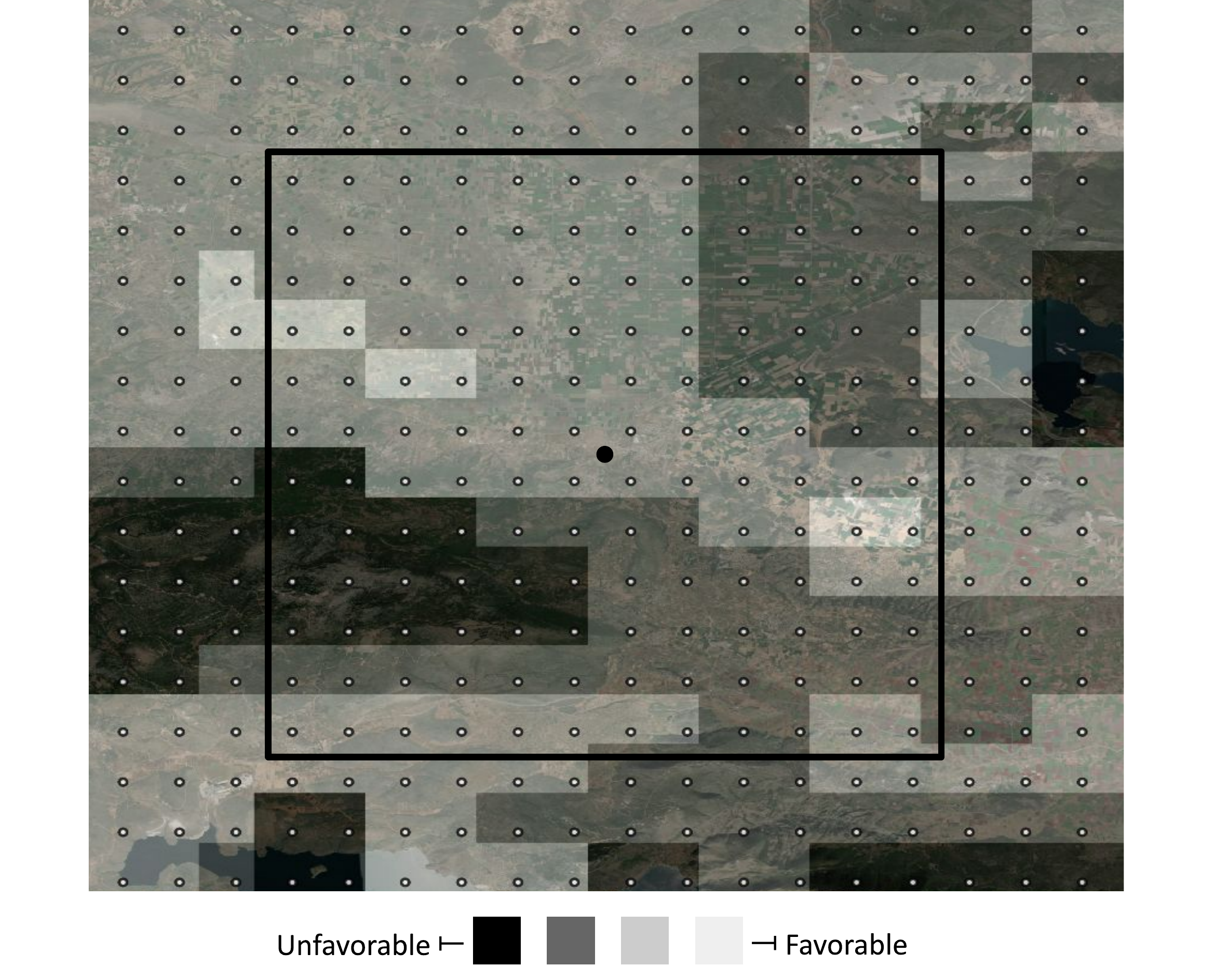}
  \caption{Optimal sowing map for a given day. The black circle at the center depicts the GFS grid point that represents the entire black-lined box. The white circles depict the 144 ART grid points for the same area.  \label{fig:spatial_gfsvsart}}
\end{figure}

% In the Figure \ref{fig:gfs_art_runs} are presented some randomly selected 10-days forecast runs for a visual and intuitive comparison between GFS and ART. Furthermore,

We generate ART forecasts in order to provide recommendations that can vary up to the field-level, which would have been impossible with GFS forecasts alone. This is depicted in Figure \ref{fig:spatial_gfsvsart}. In order to evaluate the quality of our ART forecasts, we compared them with measurements from the nearest operational weather station in the area of interest for the critical sowing period, from 15/4/2021 to 15/5/2021. We have limited our comparison to the maximum and minimum ambient temperatures, as there were not any soil temperature measurements available. It is worth noting that the nearest grid point of GFS to the station is only $0.87$ km away, however the maximum distance can be up to 12 km away. On the other hand, the equivalent grid point of ART is $1.41$ km away, which incidentally is the maximum possible distance between any location and the nearest ART point.
% Therefohe GFS error would be relatively small, but in reality this error is  performance will occur in the locations in between of $12.5$ km where only a GFS forecast is assigned to them given its medium spatial resolution.

% This issue of spatial resolution we tried to overcome with ART in order to be able to generate a near to the field-level resolution recommendations and do not lose the variability of recommendations at field-level as one can see in Figure \ref{fig:spatial_gfsvsart} .

Initially, we compared the next day forecasts of GFS against their ART (or WRF) equivalent. The comparison analysis revealed a Mean Absolute Error (MAE), between the two forecasts and the station for maximum ambient temperature, equal to $2.39^\circ$C (GFS) versus $1.48^\circ$C (ART), and for minimum ambient temperature $1.52^\circ$C (GFS) versus $1.74^\circ$C (ART). Overall, WRF appears to behave well and slightly better than GFS. This difference is expected to be greater for other locations in the grid, as for this particular case the station happened to be very close to the GFS grid point. 
Furthermore, we calculated the MAE and Root Mean Squared Error (RMSE) of all daily 5-day forecasts of ART against the ground station for a period of interest. For the maximum temperature we found $MAE=2.41$, $RMSE=3.11$, whereas for the minimum temperature we found $MAE=2.75$, $RMSE=3.70$. A graphical comparisons of ART forecasts against the ground station measurements is presented in the Figure \ref{fig:comparisons})

\begin{figure}[!ht]
  \centering
  \includegraphics[scale=0.40]{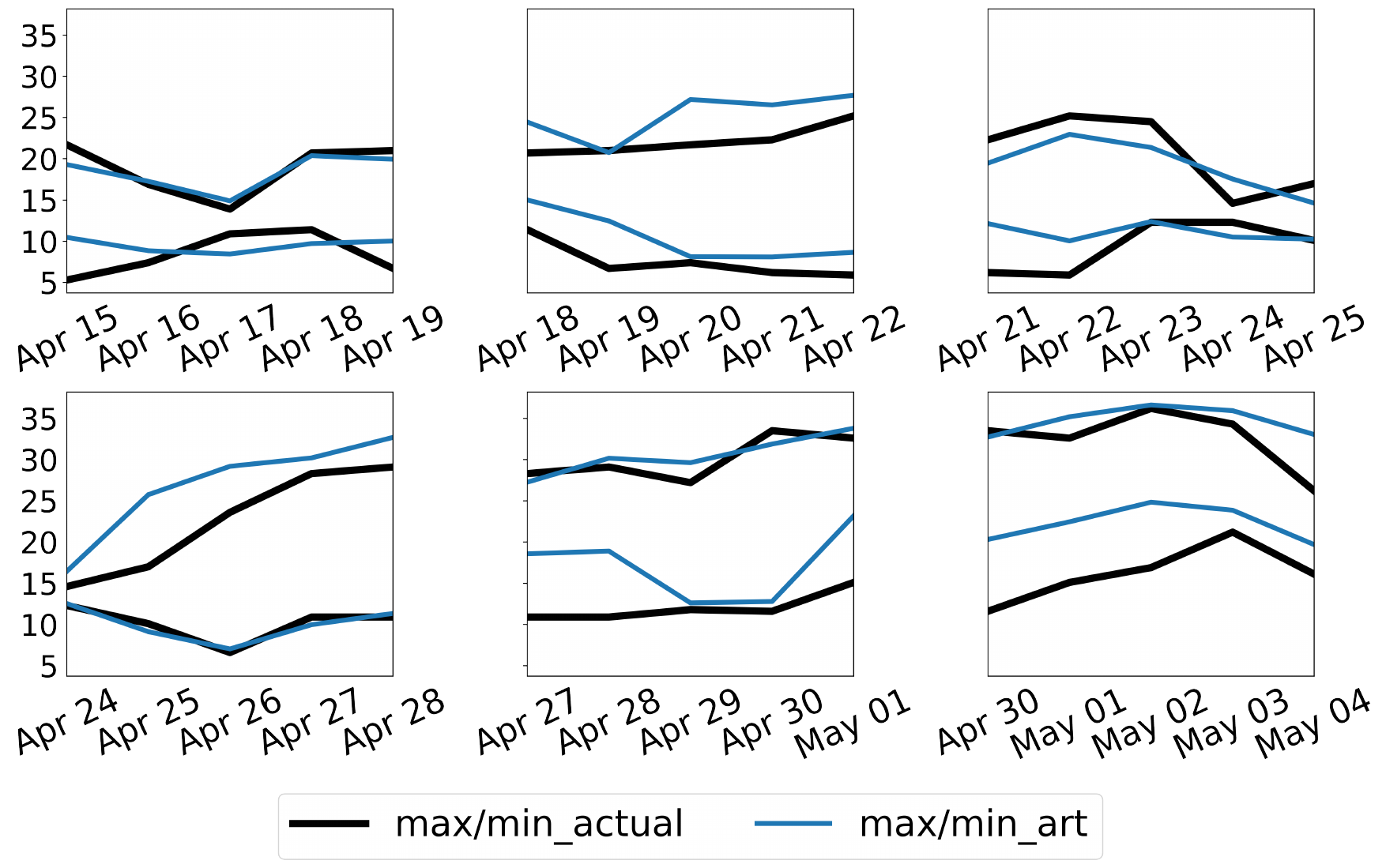}
  \caption{5-day max/min temperature for ART forecasts and ground station measurements. Each plot shows a different 5-day prediction.  \label{fig:comparisons}}
\end{figure}

\subsection{Cotton Domain Knowledge and Graph Building}

% In order to estimate whether sowing action recommendations \sout{resulted in realizing higher} affect \sout{somehow} yields, a representative causal graph of agro-environmental system that included the dictated by domain knowledge number of the relative variables was constructed.
Cotton yield and quality are ultimately determined by the interaction between the genotype, environmental conditions and management practices throughout the growing season. Nevertheless, the first pivotal steps for a profitable yield are a successful seed germination and emergence which are greatly dependent on timely sowing \cite{wanjura1969emergence, bauer1998planting,  bradow2010germination}.
% Successful cotton seed germination and emergence are the first pivotal steps for obtaining a profitable yield \cite{wanjura1969emergence}; punctual sowing action affects these steps and therefore the final yield \cite{bradow2010germination, bauer1998planting}.
% So, in this work indirect we also check and quantify the effect of this first step of a good sowing, taking account that in the labeling of a good and bad sowing is encompassed an amount of noise by forecast errors in which RS is based on.

Emergence and germination mediate the effect of $T$ on $Y$; however, Crop Growth ($CG$) was not observed. We thus turned to the popular Normalized Difference Vegetation Index (NDVI) in order to obtain a reliable proxy of $CG$, and specifically used the trapezoidal rule across NDVI values from sowing to harvest %or any season of interest (i.e. emergnece phenological stage) 
\cite{eklundh2015timesat}. Even though in the case of cotton, trapezoidal NDVI is not linearly correlated with yield \cite{dalezios2001cotton, zhao2007canopy}, it is correlated with early season Leaf Area Index (LAI) \cite{zhao2007canopy}, which in turn is a good indicator of early season crop growth rate \cite{virk2019physiological}.
Furthermore, seed germination and seedling emergence are greatly dependent on soil moisture. Hence, soil moisture $SM$ is a confounder for the relation $T \rightarrow Y$. As a $SM$ proxy, we used the well-known Normalized Difference Water Index (NDWI) at sowing day which is highly correlated with soil moisture in bare soil \cite{casamitjana2020soil}. 
% We note that the majority of cotton fields were watered by rain prior to sowing and sufficient soil moisture was ensured through standard management practices during the growing season.

Agricultural management practices before sowing ($AbS$) comprise tilling operations for preparing a good seedbed. Practices during sowing ($AdS$) include a sowing depth of $4-5$ cm and an average distance of $0.91$ m between rows and $7.62$ cm between seeds. After sowing practices ($AaS$) comprise basic fertilization, irrigation and pest management. It is reasonable to think that all aforementioned practices are a result of a common cause that we can define as Agricultural Knowledge ($AK$), capturing the skills and expertise of a farmer. We possess no quantitative information on the agricultural knowledge or the practices followed by each farmer. However, the farmer's cooperative is not large, and aims for consistent, high-quality produce. As a result, they have developed highly consolidated routines for interacting with their crops: this includes common practices, homogeneous fertilizer application, and jointly owned machinery. We thus note that even if we do not have numerical data on $AbS, AdS, AaS$, the cooperative directors do not observe significant differences across fields and for the purposes of our study these variables are considered to be constant. 

At the same time, it is rational to assume that the agricultural knowledge ($AK$) of any farmer interacts with crops exclusively through management practices. Because of the aforementioned condition, the influence of $AK$ on the system is nullified and we hence omit it from the graph. While we note that the above limit the external validity of our results \cite{calder1982concept}, by assuming that agricultural practices are constant for all farmers and that $AK$ only interacts with the system through them, we implicitly control for all of them \cite{huntington2021effect}.

Apart from soil moisture, soil and ambient temperatures at the time of sowing and for 5-10 days after, affect seed germination, seedling development and final yield \cite{virk2019physiological,boman2005soil,varcosoil}.
Low temperatures result in reduced germination, slow growth and less vigorous 
% and/or malformed
seedlings that are more prone to diseases and sensitive to weed competition \cite{christiansen1969season,wanjura1969emergence,bradow2010germination}. This knowledge is incorporated in the sowing recommendations, in the form of numerical rules, and consequently in the treatment $T$. We thus added in the graph the weather forecast $WF$ (variables listed in Table \ref{fig:ag_rules}) as a parent node of $T$. We also had access to the weather on the day of sowing $WS$ (min \& max ambient temperature in $^\circ$C) from a nearby weather station, influencing $WF$, $T$, and $CG$.

% According to previous research, the lowest minimum mean soil temperature for cotton germination has been determined to be at 16C \cite{bradow2010germination}, while soil and ambient temperatures lower than 10C result in less vigorous and malformed seedlings \cite{boman2005soil}. As a general rule, soil mean temperatures higher than 18C with a forecast of ambient temperatures higher than 26C for at least 5 days after sowing are recommended to cotton farmers \cite{boman2005soil}. 

% Since temperature is the primary environmental factor controlling plant growth \cite{bange2004impact,hatfield2015temperature} temperature fluctuations were partially observed throughout the growing season.  

Topsoil (0-20 cm) properties $SP$ (\% content of clay, silt and sand) and organic carbon content  $SoC$ (g C kg\textsuperscript{-1}) also affect cotton seed germination and seedling emergence due to differences in water holding capacity and consequently in soil temperature and aeration, drainage and seed-to-soil contact \cite{varcosoil}. Data on $SP$ and $SoC$ were retrieved from the European Soil Data Centre (ESDAC) \cite{ballabio2016mapping, de2015map}. Both variables were included in the graph as confounders of $T$ and $CG$.
Seed variety also determines seed germination, emergence and final yield \cite{sniderseed}.
% as significant genotypic variability in terms of seedling vigor has been observed among cotton genotypes
% Seedling vigor is a poorly defined term but in general refers to the competitive ability of the plants at the seedling stage with high vigor seedlings being less sensitive to early season pests and diseases, more competitive against weeds and able to establish a good plant stand (Pilon et al. 2016). 
Seed mass and vigor \cite{liu2015early,sniderseed} are related to the seed variety ($SV$); we hence added the latter as a confounder for $T$ and $Y$. In this case, we had 13 different cotton SVs.

\begin{table}[!ht]
\small
\centering
% \resizebox{\columnwidth}{!}{%
\begin{tabular}{lll}
\toprule
\textbf{Id} & \textbf{Variable Description}           & \textbf{Source}         \\ \midrule
T   & Treatment & Recommendation System      
% T   & Treatment & Recommender System    
\\
% Y   & Outcome (Yield) & Agriculture Cooperative \\
WF  & Weather forecast             & GFS, WRF                \\
WS  & Weather on sowing day        & Nearest weather station \\
WaS & Weather after sowing         & Nearest weather station                  \\
CG  & Crop Growth                  & NDVI via Sentinel-2                      \\
SM  & Soil Moisture on sowing      & NDWI via Sentinel-2     \\
SP  & Topsoil physical properties  & Map by ESDAC \\
SoC & Topsoil organic carbon       & Map by ESDAC \\
SV  & Seed Variety                 & Farmers' Cooperative \\
G   & Geometry of field           & Farmers' Cooperative                  \\
AdS & Practices during sowing      & Farmers' Cooperative \\
AbS & Practices before sowing      & Farmers' Cooperative \\
AaS & Practices after sowing       & Farmers' Cooperative \\
% LS  & Length of Season             & Agriculture Cooperative \\
HD  & Harvest Date                 & Farmers' Cooperative \\
Y   & Outcome (Yield)              & Farmers' Cooperative \\ \bottomrule
\end{tabular}%
\caption{Farm system variable identifier, description and source.}
\label{tab:variables_sources}
% }
\end{table}

The geometrical properties of the field (perimeter to area ratio, $G$) were also considered, as border effects can play a minor role on crop growth, confounding the effect of $T$ on $Y$
% due to \cite{green1956border} and sowing procedure. 
\cite{green1956border}. Since temperature is the primary environmental factor controlling plant growth \cite{bange2004impact,hatfield2015temperature}, temperature fluctuations were observed throughout the growing season from the nearest weather station, constituting a parent variable $WaS$ (min \& max ambient temperature in $^\circ$C) of crop growth $CG$. 
% Lastly, the length of growing season (days from sowing till harvest, $LS$), 
% and intrinsically controlled by plant growth consequently and the temperature was one more variable that was included in the causal graph coupled with
Lastly, the Harvest Date ($HD$) mediates the effect of $CG$ on $Y$, influencing both yield potential and quality \cite{dong2006yield,bange2008managing}. Table \ref{tab:variables_sources} summarizes the variables' description, abbreviation and source.  
% Specifically, warm temperatures early in the season allow early sowing, establishment of a good plant stand and enough time for maturity while cold temperatures at the end of the season ensure good timing of crop maturity and effectiveness of chemical harvest aids which both affect cotton yield and quality \cite{bange2004impact,stiller2004maturity}.

\subsection{Implementation \& Results Details}

% dowhy and causalml
For the experiments, we are using the popular doWhy \cite{sharma2020dowhy} and Causal ML \cite{chen2020causalml} Python libraries.
 
 % propensity score and balance
Propensity modeling is a prerequisite of IPS weighting. We thus begin by discussing the propensity model that is fit. Given the relatively small dataset size, 
% and the fact that we are not optimizing for predictive performance (cite kiciman?),
logistic regression is used on the scaled back-door adjustment set $Z$ for classifying each field into the treatment/control group. We subsequently trim the dataset by removing all rows with extreme propensity scores ($<0.2$ or $>0.8$) to aid the overlap assumption
% that is fundamental for matching estimators 
\cite{imbens2015causal}. The resulting distribution of propensity scores can be seen at Figure \ref{fig:prop_distr_trimmed}. The model scores $0.81$ in accuracy, $0.64$ in F1-score, and $0.88$ in ROC-AUC.
% while evaluation metrics are included in Table \ref{tab:prop_performance}. 
After trimming extreme propensity scores, a subset of $48$ treated and $37$ control units remains. There is decent overlap between the propensity score distributions of the treatment and control group, indicating that they are comparable and enabling reliable propensity-based ATE estimation.

\begin{figure}[!ht]
  \centering
  \includegraphics[scale=0.23]{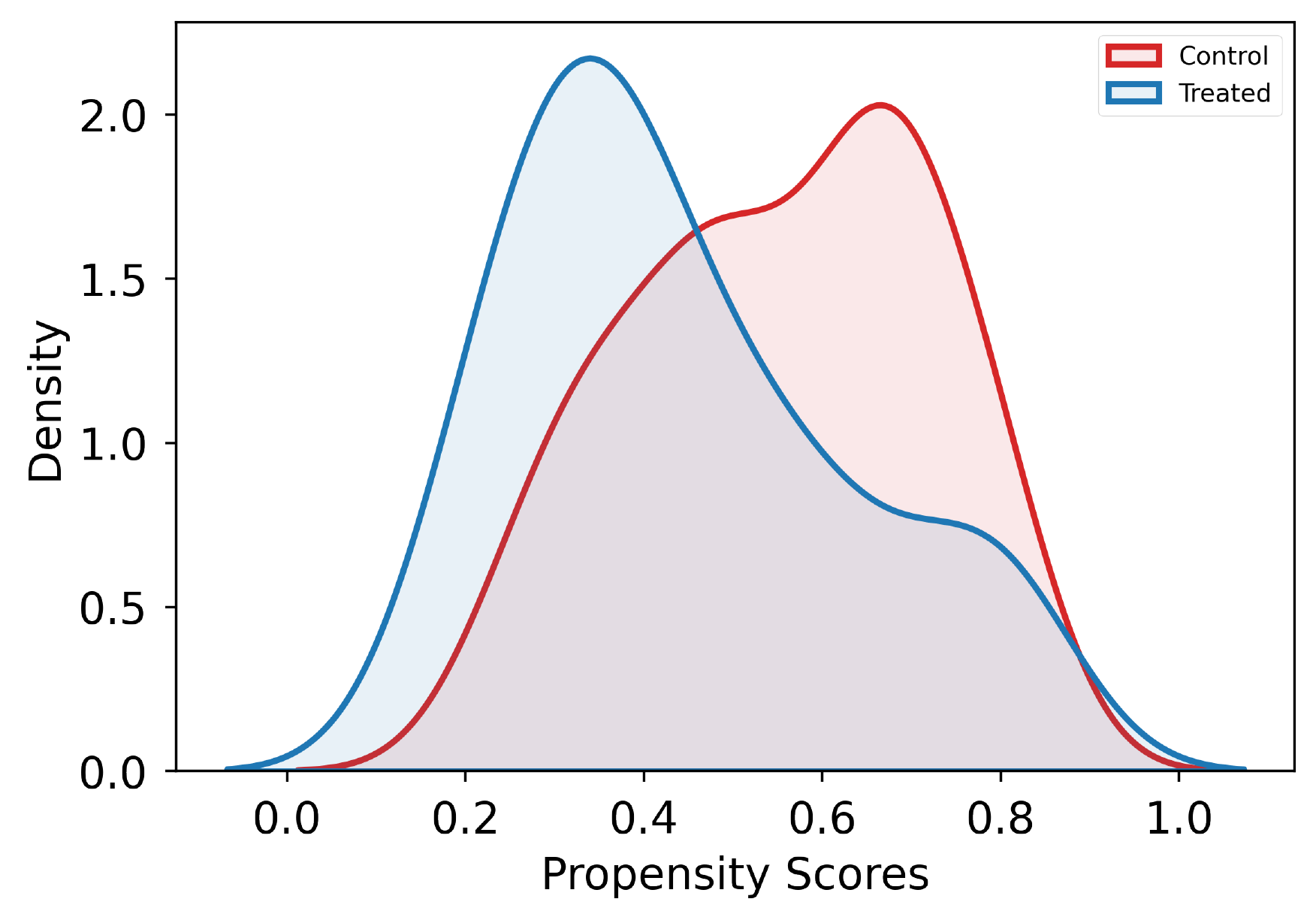}
  \caption{Distribution of propensity scores for the control and treatment group after trimming extreme scores.} \label{fig:prop_distr_trimmed}
  %\vspace{-2em}
\end{figure}

% bootstraping/refutation 1000 times
Besides Linear Regression, other methods do not provide confidence intervals by default. For matching, IPS, and meta-learners confidence intervals and the resulting p-values are hence bootstrapped. Also, the Placebo, RCC and RSR refutations tests are bootstrapped to generate confidence intervals and p-values \cite{diciccio1996bootstrap}. Confidence intervals and p-values are bootstrapped ($1000$ iterations). 

The UCC refutation test returns a heatmap of new ATE estimates depending on the strength of injected unobserved confounding. Figure \ref{fig:example} contains heatmaps with the impact of unobserved confounding on the ATE estimate for the linear regression, matching, and IPS weighting methods which were deployed through doWhy. Observing the heatmaps, we note that the estimation methods are robust to a moderate amount of unobserved confounding, in the sense that the ATE values of the lower region of each heatmap (where the effect of the unobserved confounder does not dominate the treatment and outcome values) largely remains positive and comparable to the real ATE estimate. We note that as the strength of unobserved confounding increases, significant volatility in effect estimates is expected, as the effect is no longer fully identified. For more information about the implementation and the proper interpretation of test, see the doWhy library documentation \cite{sharma2020dowhy}.

\begin{figure}[!ht]
    \centering
    \subfloat{{\includegraphics[scale=0.25]{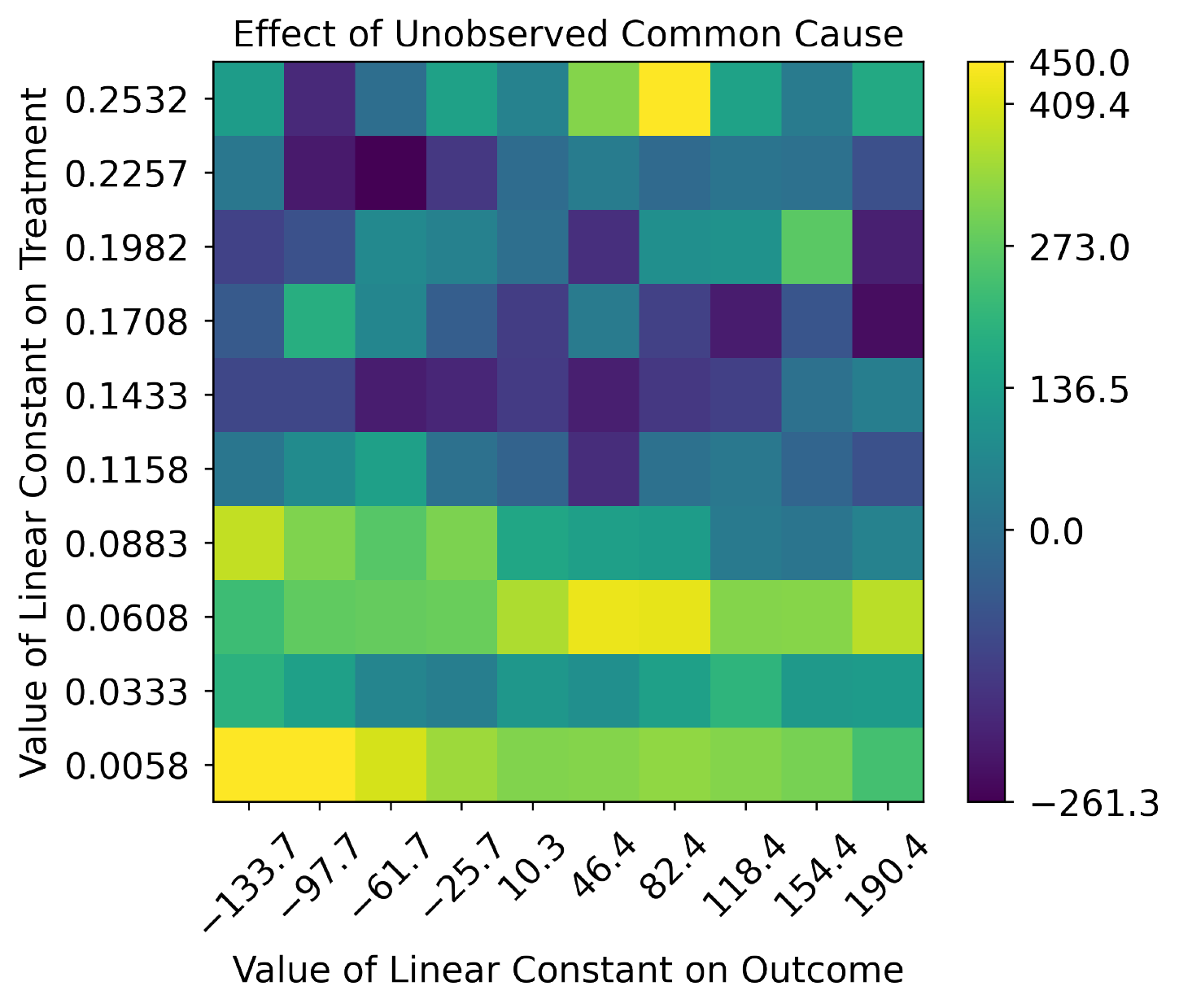} }}
    \qquad
    \subfloat{{\includegraphics[scale=0.25]{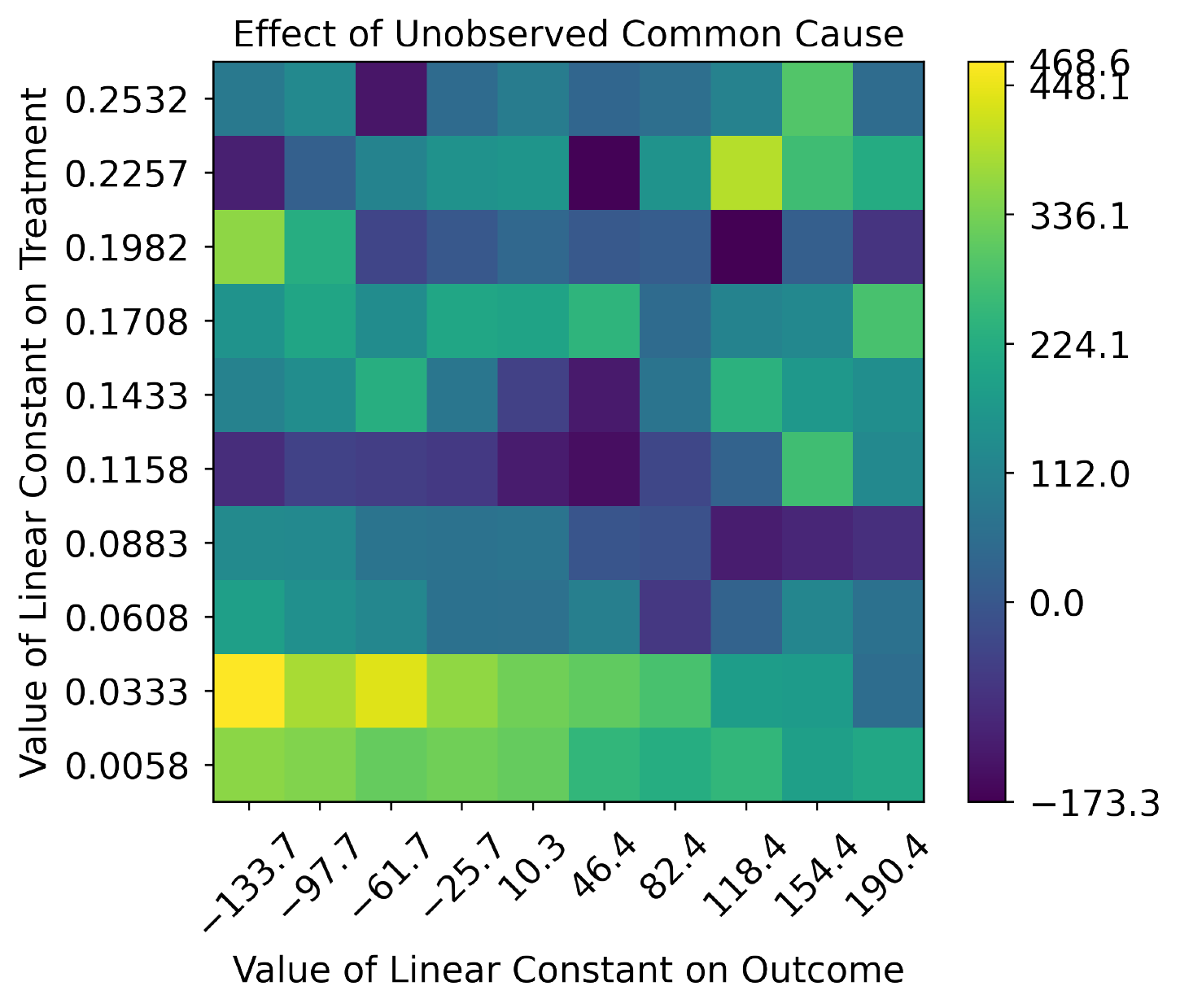} }}
    \qquad
    \subfloat{{\includegraphics[scale=0.25]{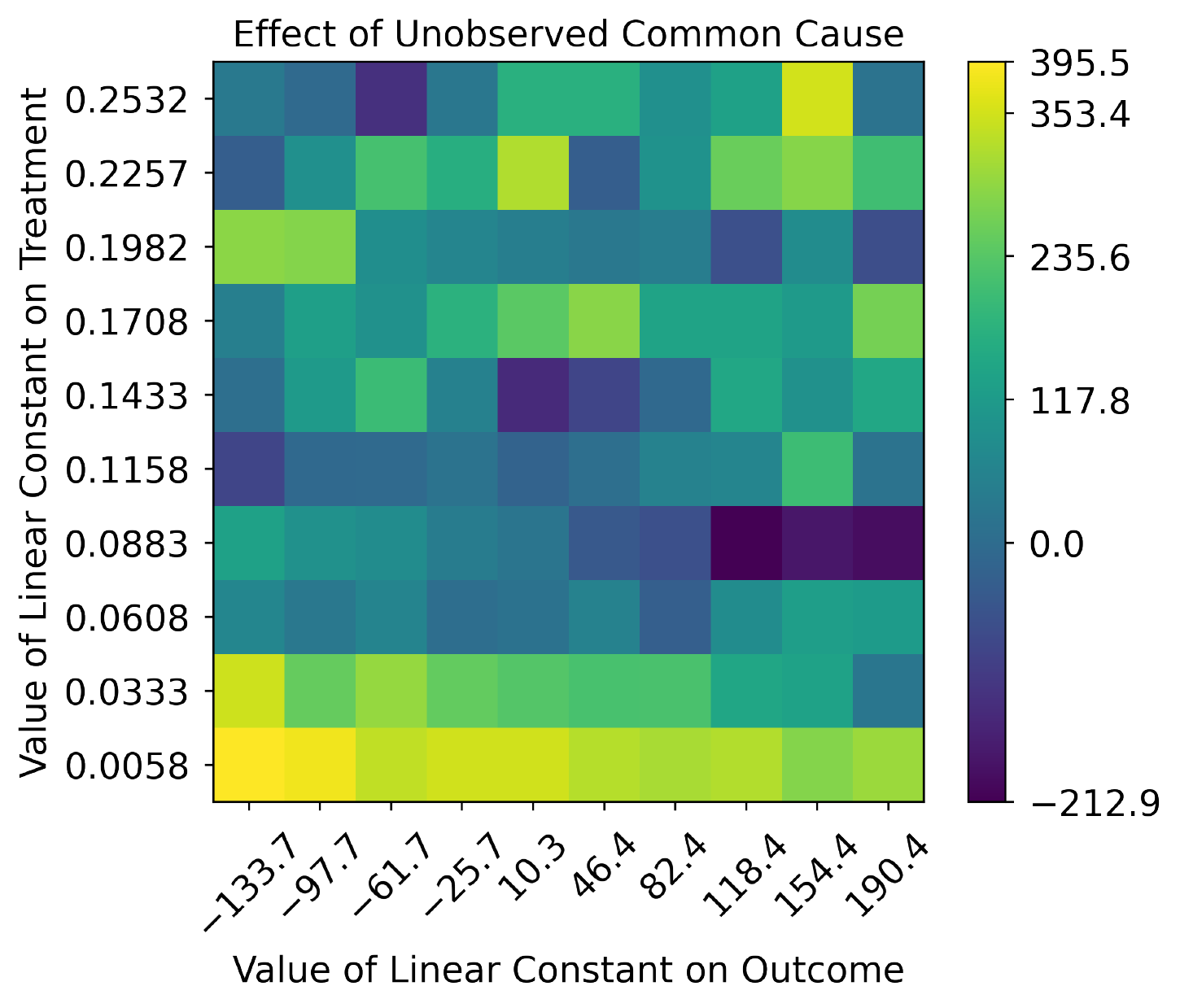} }}
    \caption{Unobserved Common Cause heatmap results for Linear Regression (Top Left), Matching (Top Right) and IPS weighting (Bottom) In the main paper, we report the average cell value of each heatmap (i.e, the average ATE across multiple combinations of unobserved confounding.)}
    \label{fig:example}
\end{figure}

}
\end{document}